%% file: main.tex
\documentclass[letterpaper, 10 pt, conference]{ieeeconf}  %

\IEEEoverridecommandlockouts                              %

\usepackage{graphics} %
\usepackage{epsfig} %
\usepackage{mathptmx} %
\usepackage{times} %
\usepackage{amsmath} %
\usepackage{amssymb}  %
\usepackage{lineno}
\usepackage{cite}
\usepackage[breaklinks, colorlinks=true, linkcolor=blue, citecolor=blue, urlcolor=blue]{hyperref}

\usepackage[capitalize,nameinlink]{cleveref}
\crefname{section}{Sec.}{Secs.}
\Crefname{section}{Section}{Sections}
\Crefname{subsection}{Subsection}{Subsections}
\crefname{subsection}{Sec.}{Secs.}
\Crefname{table}{Table}{Tables}
\crefname{table}{Tab.}{Tabs.}
\Crefname{figure}{Figure}{Figures}
\crefname{figure}{Fig.}{Figs.}
\Crefname{equation}{Equation}{Equations}
\crefname{equation}{Eq.}{Eqs.}

\newcommand{\eg}{\textit{e.g.}}
\newcommand{\ie}{\textit{i.e.}}

\input{preamble}

\newcommand{\para}[1]{\vspace{0.5em}\noindent\textbf{#1}}

\showeditsfalse

\title{\LARGE \bf
\titletext
}

\author{Tetiana Martyniuk$^{1,2}$ \quad Gilles Puy$^{1,2}$ \quad Alexandre Boulch$^{1,2}$ \quad Renaud Marlet$^{1,2,3}$ \quad Raoul de Charette$^{1}$%
\thanks{$^{1}$ {Inria}, $^{2}$ {valeo.ai}, $^{3}$ {LIGM, ENPC, Univ Gustave Eiffel, CNRS, France}}%
}

\begin{document}

\maketitle
\thispagestyle{empty}
\pagestyle{empty}

\input{sections/00_abstract}

\input{sections/01_intro}

\input{sections/02_related}

\input{sections/03_method}

\input{sections/04_results}

\input{sections/05_conclusion}

\input{sections/0X_appendix}

\bibliographystyle{IEEEtran}
\bibliography{main}

\end{document}

%% file: preamble.tex
\usepackage{pifont}
\usepackage[dvipsnames]{xcolor}
\usepackage{graphicx} %
\usepackage{amsmath}
\usepackage{amssymb}
\usepackage{adjustbox}
\usepackage{multirow}
\usepackage{colortbl}
\usepackage{cleveref}
\usepackage{url}
\usepackage{booktabs}

\newif\ifshowedits

\newcommand{\addeditor}[3]{%
  \definecolor{#1color}{rgb}{#3}
  \expandafter\newcommand\csname #1\endcsname[1]{%
  \ifshowedits
    {\color{#1color} ##1}%
  \else
    {##1}%
  \fi
  }%
  \expandafter\newcommand\csname #1rmk\endcsname[1]{%
  \ifshowedits
    {\color{#1color} {\bf [#2: ##1]}}
  \fi
  }%
  \expandafter\newcommand\csname #1rpl\endcsname[2]{%
  \ifshowedits
    {\color{#1color} ##1 \sout{##2}}
  \else
    {##1}
  \fi
  }%
}

\definecolor{MyGreen}{RGB}{0, 180, 0}
\definecolor{MyRed}{RGB}{180, 0, 0}
\definecolor{MyYellow}{RGB}{180, 180, 0}

\newcommand{\titletext}{{LiDPM: Rethinking Point Diffusion for Lidar Scene Completion}}
\newcommand{\method}{LiDPM}
\newcommand{\ddpm}{DDPM}

%% file: sections/00_abstract.tex
\begin{abstract}
Training diffusion models that work directly on lidar points at the scale of outdoor scenes is challenging due to the difficulty of generating fine-grained details from white noise over a broad field of view. 
The latest works addressing scene completion with diffusion models tackle this problem by reformulating the original \ddpm{} as a local diffusion process. 
It contrasts with the common practice of operating at the level of objects, where vanilla \ddpm{}s are currently used. 
In this work, we close the gap between these two lines of work. 
We identify approximations in the local diffusion formulation, show that they are not required to operate at the scene level, and that a vanilla \ddpm{} with a well-chosen starting point is enough for completion. 
Finally, we demonstrate that our method, \method{}, leads to better results in scene completion on SemanticKITTI.
The project page is \url{https://astra-vision.github.io/LiDPM}~.
\end{abstract}

%% file: sections/01_intro.tex
\section{Introduction}

Lidars are key sensors for autonomous driving, measuring accurate distances to the vehicle environment. %
However, lidar point clouds are sparse, leaving wide gaps between scanned points. 
Yet, filling these gaps benefits downstream tasks like mapping~\cite{popovic2021volumetric,vizzo2022makeitdense} or object detection~\cite{wu2022sparse,xiong2023ultralidar,shan2023scp}.
Besides, acquisition patterns vary with sensor models (\eg, beam count) and placement, affecting the transferability of perception algorithms (\eg, detection, segmentation).
There are several ways to tackle these domain gaps, such as domain adaptation~\cite{michele2024saluda}, domain generalization~\cite{sanchez2023domaingeneralization}, or scene completion~\cite{yi2021completeandlabel}, the latter allowing %
simulating new sensors by resampling the completed scene using different scan patterns.
Moreover, 
beyond completion, generating entirely new scenes %
can be used to create or augment datasets. %

Diffusion for point clouds offers an attractive solution for both completion and generation.
In particular, direct diffusion on points, \ie, moving points in 3D space without resorting to a latent scene encoding, %
has shown promising results~\cite{lidiff}. 
One of the main advantages is that it inherently generates points, in contrast to approaches based on voxels (discretizing space) or surfaces (requiring reconstructions prior to training).

Using denoising diffusion probabilistic models (\ddpm{}s)~\cite{ddpm} as a basis, point diffusion has been mainly studied on small shapes ($<$10K %
pts)~\cite{pvd,dit3d}, with architectures that hardly scale to large point clouds. To scale to automotive scenes ($>$100K %
pts), LiDiff~\cite{lidiff} reformulated the problem as a local diffusion process. However, it then only allows scene completion, not generation. 
Besides, this local formulation introduces unnecessary approximations.

In this paper, we propose a new method, called \method{}, that brings together these two perspectives, extending \ddpm{}s to scenes.
Our contributions are as follows:
\begin{itemize}
    \item 
    We analyze the approximations and limitations of formulating the problem as a local diffusion.
    \item
    We show that a vanilla DDPM can be used for a global diffusion on large point clouds, thus unifying point diffusion frameworks for shapes (objects) and scenes.
    \item We demonstrate that \method{} outperforms local diffusion for scene completion on SemanticKITTI~\cite{semantickitti}. \cref{fig:teaser} illustrates the quality of our results.
\end{itemize}

\input{figures/teaser_no_refine}

%% file: figures/teaser_no_refine.tex
\begin{figure}
    \centering
    {%
        \begin{tabular}{c}
        \small (a) \method{}$^\dagger$ (ours) w/o refinement, implementing \emph{global} diffusion
        \\[1mm]
        \includegraphics[width=0.8\linewidth, trim={0 2cm 0 0},clip]{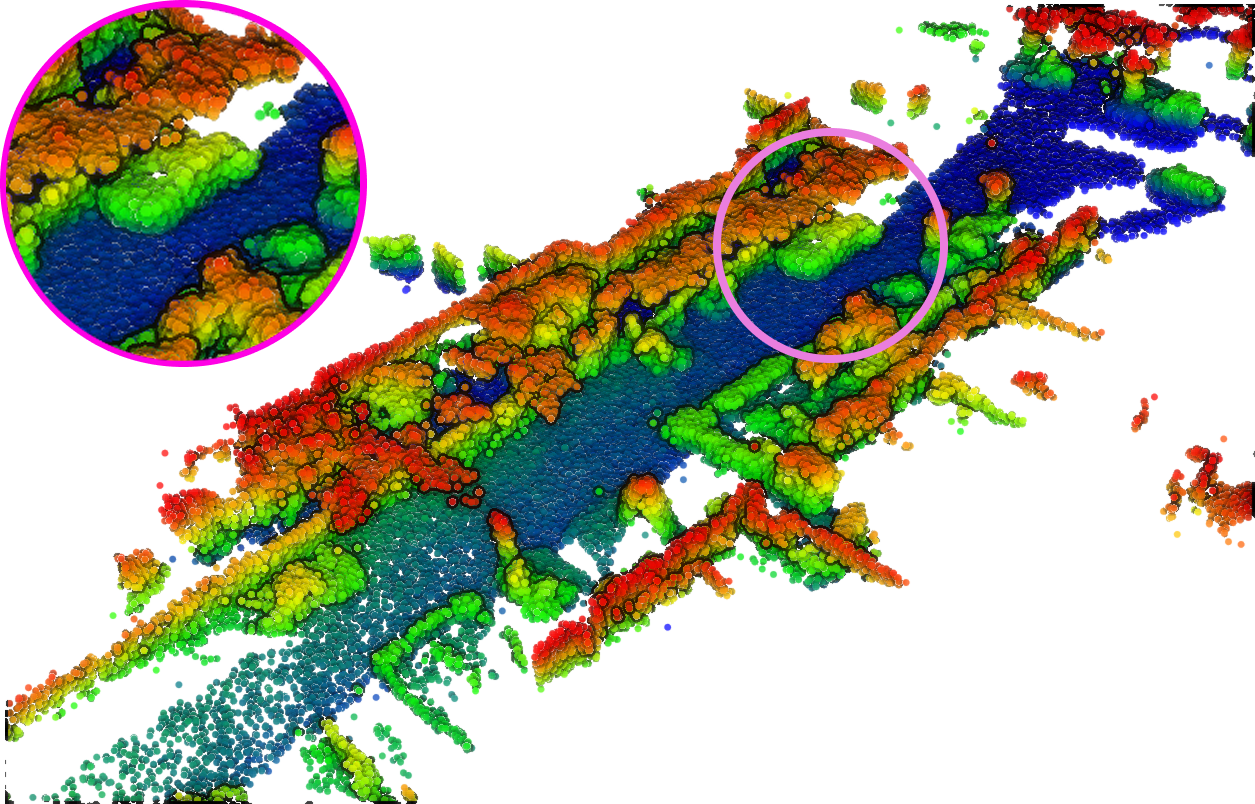} %
        \\[1mm]
        \small (b) LiDiff$^\dagger$~\cite{lidiff} w/o refinement, implementing \emph{local} diffusion
        \\[1mm]
        \includegraphics[width=0.8\linewidth, trim={0 2cm 0 0},clip]{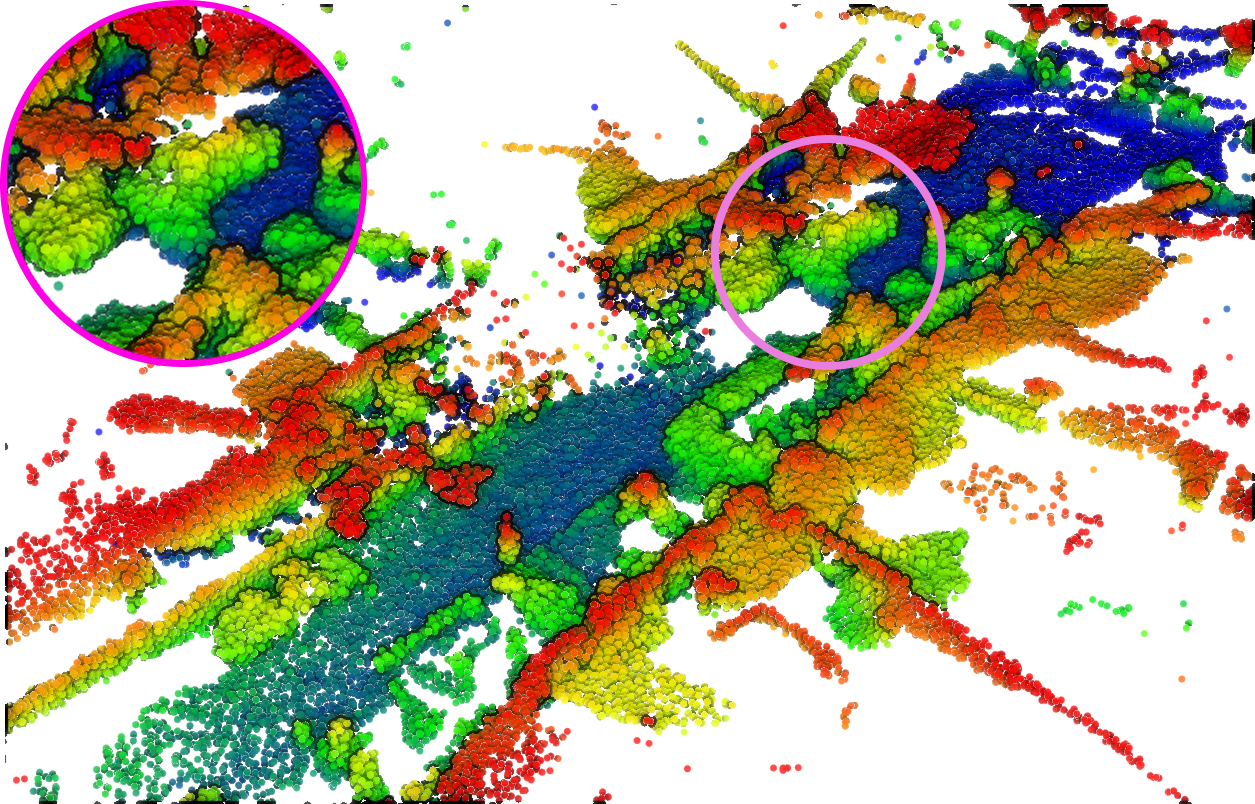} %
        \\
        \vspace*{-6mm}
        \end{tabular}
    }
    \caption{
    \textbf{Scene-level point diffusion for completion}. 
    Our \method{}$^\dagger$ formulation (top) follows the general \ddpm~paradigm, yielding more realistic and accurate completions than LiDiff$^\dagger$ local diffusion~\cite{lidiff} (bottom).
    }
\vspace*{-3mm}
    \label{fig:teaser}
\end{figure}

%% file: sections/02_related.tex
\section{Related work}
\label{sec:related}

\input{figures/methods}

\subsection{Diffusion on points}

Over the past years, diffusion has been applied to various generative tasks, from image or video synthesis to speech generation. 
Most methods are based on \ddpm{}s~\cite{ddpm}, 
which define a forward process that gradually adds Gaussian noise to the input, and a reverse process that learns to recover the input from the noise.

{
Methods that apply \ddpm{}s directly to the point clouds~\cite{pvd, luo2021diffusion, dit3d} operate by moving points in 3D space, {addressing generation or completion at the point level}.
In PVD~\cite{pvd}, the conditioning for completion (concatenating sparse and noisy point cloud) imposes separate training for completion and generation (\cref{fig:methods}, top).
Furthermore, all the above methods focus on shapes (objects) and do not {naturally} scale to scene-level generation and completion.
}

Recently, LiDiff~\cite{lidiff} proposed scaling the diffusion process to large scenes to complete automotive lidar scans.
The authors scale up the backbone and argue for a novel ``local diffusion'' formulation.
By doing so, LiDiff can complete scenes but at the cost of approximations, which (i)~prevent using the same formulation for generation~(\cref{fig:methods}, middle) and (ii)~require additional regularization to stabilize training.
We study these limitations in the \cref{app:localdiff}\!\!. %

In comparison, we show that we can unify the practices between the different scales (shapes/objects and scenes).
Instead of a custom diffusion, we build on the original DDPM formulation (closer to the current practice for shapes), scale the backbone to allow scene completion (from a dense to a sparse CNN), and use conditioning to allow both completion and generation (\cref{fig:methods}, bottom).

\subsection{{Scene Completion}}
{Historically, scene completion was first achieved by completing depth or lidar data in the image space, typically relying on custom strategies to train CNNs~\cite{ma2018sparse,jaritz2018sparse}. 
While these methods could benefit from lightweight 2D networks, they were also inherently limited to visible areas within the field of view. 
Therefore, a large body of works now addresses the scene completion task in the 3D space for denser completions, also fueled by the emergence of SemanticKITTI~\cite{semantickitti}, which first provided a benchmark by aggregating lidar scans.%
}
{%
MID~\cite{vizzo2022makeitdense} trains a network that, given a single scan, estimates a signed distance field from voxels to the complete scene. 
The actual scene completion is then obtained via the Marching Cubes algorithm~\cite{lorensen1987marchingcubes}. 
LODE~\cite{li2023lode} also operates at the voxel level, estimating voxel occupancy. 
Aside from our main objective, some methods also leverage and infer semantics~\cite{roldao2020lmscnet, scpnet, sscnet, yan2021sparse} and are surveyed in the work of Roldão et al.~\cite{roldao20223d}.
However, as they operate primarily on voxels, these methods offer accuracy limited to the voxel resolution. 
A particular case is Local-DIFs~\cite{rist2022semscenecomp}, which learns an implicit function that can be queried at arbitrary positions.}

\para{Point-based scene completion.} 
Closer to us, some methods operate directly on point clouds, but mainly focus %
on object completion~\cite{cai2020shapegf,yuan2018pcn,boulch2021needrop}, with few addressing large-scale point clouds like automotive lidar scans~\cite{sulzer2022deep}. %
Typically, learning-based surface reconstruction algorithms train on meshes or very dense point clouds, relying on surface- or point-oriented normals.
{They implicitly estimate density~\cite{cai2020shapegf} or occupancy~\cite{boulch2021needrop} in continuous space but do not scale to scenes and still suffer from voxel-like discretization when reconstructing surfaces via the Marching Cubes algorithm.}
Some works tackle scene-level surface reconstruction~\cite{vizzo2021poisson,huang2023nksr}, but unlike our approach, they require dense input point clouds at inference, \eg, aggregated scans; in contrast, we only use a single sparse lidar scan.
Few methods specifically address point-based scene completion with diffusion models, such as LiDiff, as previously discussed.

%% file: figures/methods.tex
\begin{figure}
    \centering
    \includegraphics[width=1.0\linewidth]{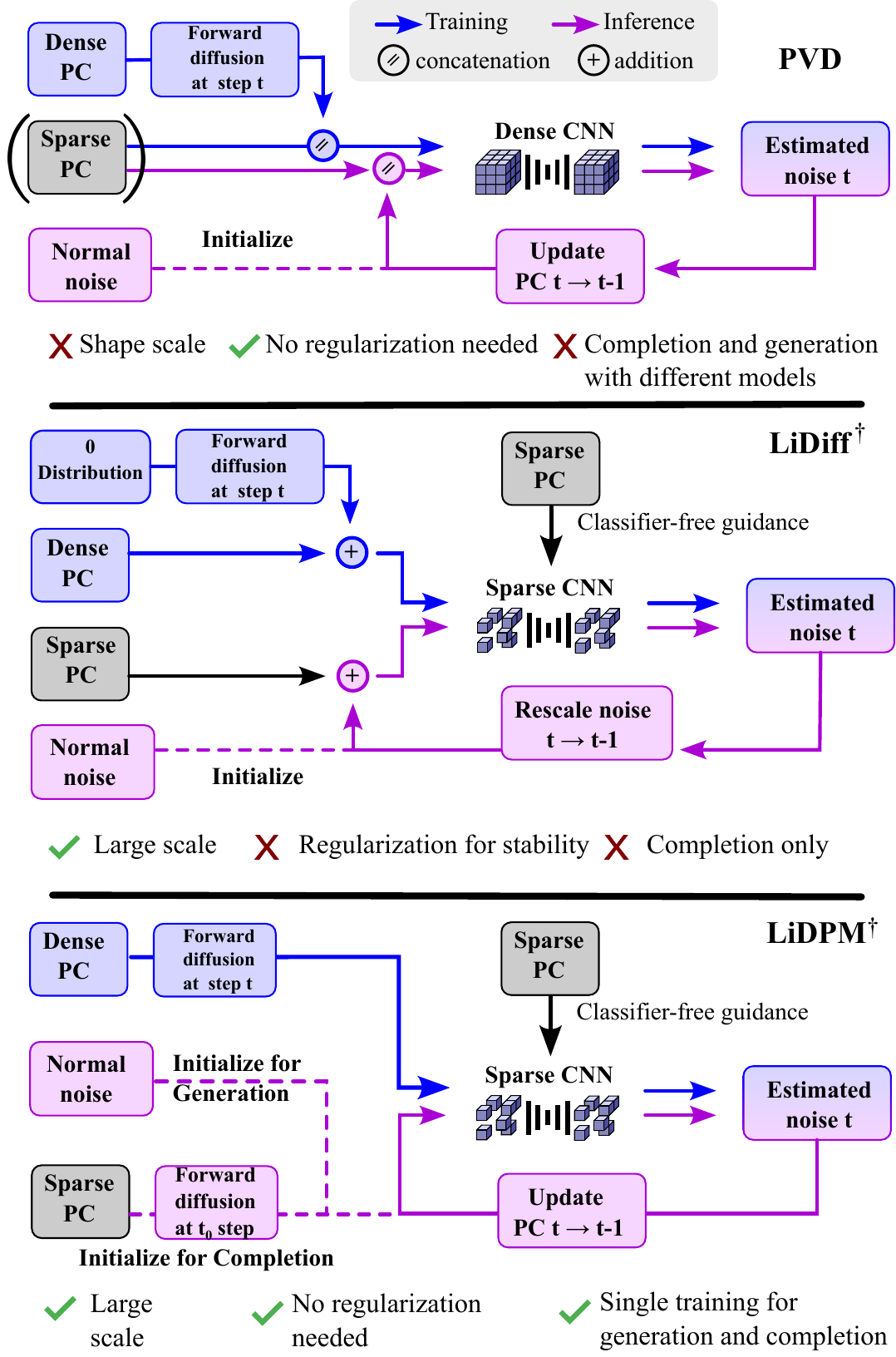}
    \vspace*{-3mm}
    \caption{\textbf{Point diffusion methods.} PVD~\cite{pvd} (top) applies \ddpm{} at the object level. LiDiff~\cite{lidiff} (middle) uses local diffusion to work at the scene level. \method{}~(bottom) makes it possible to use \ddpm{} at the scene level.}
    \label{fig:methods}
\vspace*{-5mm}
\end{figure}

%% file: sections/03_method.tex
\section{Method}
\label{sec:method}

\method{} addresses the task of pointwise scene completion. 
It aims at predicting a dense point cloud~$\mathbf{p}^d$ from a sparse point cloud~$\mathbf{p}^s$, typically a lidar scan. 
Beyond up\-sampling, this task is generative in that~$\mathbf{p}^s$ may suffer from large holes due to scene occlusions or sensor limitations, requiring synthesizing missing structures. %
To that end, we formulate the task as a denoising problem, which aims to reconstruct $\mathbf{p}^d$ conditioned on the sparse input $\mathbf{p}^s$. 

LiDiff~\cite{lidiff} followed by subsequent works~\cite{scorelidar, diffssc} argues
that the \ddpm{} formulation~\cite{ddpm} cannot be applied to large-scale lidar point clouds due to variations along the point cloud axes that result in a loss of details in the denoised reconstruction.
Thus, they reformulate the diffusion process as \emph{local} point denoising.
Different from %
these works, we show that a careful formulation of the point diffusion process and relevant design choices allow an extension of \ddpm{} to point cloud completion of large-scale urban scenes. 
Our diffusion process is formulated using \emph{global} point denoising.

In this section, we first present preliminaries about
\ddpm{}s and their use for data generation (\cref{subsec:ddpm}). %
In \cref{sec:meth_meth}, we detail our method, coined \method, which extends DDPM to lidar point cloud completion.

\subsection{Preliminaries on Diffusion Models}
\label{subsec:ddpm}
DDPMs~\cite{ddpm} are {generative models} that iteratively transform data into Gaussian noise (forward process) and learn to reverse this process through a Markov chain of denoising steps (reverse process), possibly conditioned by another signal, such as a lidar scan for scene completion.

\para{The forward diffusion process} gradually corrupts a data point $\mathbf{x}_0$, \eg, a complete point cloud, over $T$ discrete time steps, by adding Gaussian noise:%
\begin{align}
\label{eq:forward}
\mathbf{x}_t 
= 
\sqrt{1 - \beta_t} \, \mathbf{x}_{t-1} + \sqrt{\beta_t} \, \mathbf{\epsilon}_t
= 
\sqrt{\bar{\alpha}_t} \, \mathbf{x}_{0} + \sqrt{1 - \bar{\alpha}_t} \, \mathbf{\epsilon}_1,
\end{align}
where $\beta_t > 0$ is a variance schedule controlling the noise level at each step $0<t\le T$, $\mathbf{\epsilon}_t \sim \mathcal{N}(\mathbf{0}, \mathbf{I})$, and $\bar{\alpha}_t = \prod_{i=1}^{t} (1 -\beta_i)$.
For a large enough $T$, $\mathbf{x}_T$ is nearly indistinguishable from the Gaussian noise $\mathcal{N}(\mathbf{0}, \mathbf{I})$.

\para{The reverse diffusion process} seeks to sample from the original data distribution by starting from $\mathbf{x}_T \sim \mathcal{N}(\mathbf{0}, \mathbf{I})$. Sampling can be done iteratively applying the following formula from $t=T$ to $t=1$:
\begin{align}
\label{eq:denoising}
\mathbf{x}_{t-1} = \frac{1}{\sqrt{\alpha_t}}\left(\mathbf{x}_{t} - \frac{1 - \alpha_t}{\sqrt{1 - \bar{\alpha}_t}} \epsilon_\theta(\mathbf{x}_t, t) \right) + \sqrt{\beta_t} \; \mathbf{z}_t,
\end{align}
where $\alpha_t = 1 - \beta_t$ and $\mathbf{z}_t \sim \mathcal{N}(\mathbf{0}, \mathbf{I})$ if $t > 1$ or $\mathbf{z}_1=\mathbf{0}$ otherwise.
The network $\epsilon_\theta$ in \cref{eq:denoising} is trained by minimizing
\begin{align}
\label{eq:learned_neural_net}
\mathcal{L}(\theta) 
& = 
\mathbb{E}_{\mathbf{x}_0, \mathbf{\epsilon}}
\| \mathbf{\epsilon} - \epsilon_\theta(\mathbf{x}_t, t) \|^2 \nonumber =\\
& = 
\mathbb{E}_{\mathbf{x}_0, \mathbf{\epsilon}}
\| \mathbf{\epsilon} - \epsilon_\theta(\sqrt{\bar{\alpha}_t} \; \mathbf{x}_{0} + \sqrt{1 - \bar{\alpha}_t} \; \mathbf{\epsilon}, t) \|^2.
\end{align}

\subsection{\method}
\label{sec:meth_meth}

Building on this diffusion process, we propose to solve scene completion by learning a diffusion model generating dense point clouds $\mathbf{p}^d$ conditioned on sparse lidar scans~$\mathbf{p}^s$.

Contrary to the local point denoising methods~\cite{lidiff,scorelidar}, we show that the original diffusion process can be left intact assuming only one approximation on the starting point of the diffusion, leading to a simple and efficient formulation of scene completion.%

\para{Diffusion on points.} 
In our work, the forward process satisfies \cref{eq:forward} with $\mathbf{x}_0 = \mathbf{p}^d$, denoted as $\mathbf{p}_0^d$ to emphasize the step $0$. 
The network is trained using \cref{eq:learned_neural_net}. %
The reverse denoising process satisfies \cref{eq:denoising} where our sole approximation is on its starting point. 
While it is possible to start from $\mathbf{p}_T^d \sim \mathcal{N}(\mathbf{0}, \mathbf{I})$, we propose to begin the diffusion process from an intermediate step $0 < t_0 < T$. 
Even though computing the starting point $\mathbf{p}_{t_0}^d$ requires knowledge of $\mathbf{p}^d_0$, we show experimentally that good results are obtained by starting from
\begin{align}
\label{eq:start_our_diff}
\tilde{\mathbf{p}}_{t_0}^d
= 
\sqrt{\bar{\alpha}_{t_0}} \, \tilde{\mathbf{p}}^{s} + \sqrt{1 - \bar{\alpha}_{t_0}} \, \mathbf{z},  
\end{align}
where $\tilde{\mathbf{p}}^s$ is obtained by duplicating $K$ times the preprocessed point cloud $\mathbf{p}^s$ (see \cref{sec:results}, Implementation details), and $\mathbf{z} \sim \mathcal{N}(\mathbf{0}, \mathbf{I})$. 
The factor $K$ is chosen so that the number of points in $\tilde{\mathbf{p}}^s$ and preprocessed $\mathbf{p}^d$ have the same order of magnitude (typically $K=10$, as in LiDiff).

\para{Conditioning on $\mathbf{p}^s$.} 
We use classifier-free guidance~\cite{classifierfree} to ensure that the details captured in $\mathbf{p}^s$ are preserved in the generated point cloud.
In practice, the update $\epsilon_\theta(\mathbf{p}^d_t, t)$ in the reverse denoising process is replaced by
\begin{align}
\label{eq:our_cfg}
\tilde{\epsilon}_\theta(\mathbf{p}^d_t, t, \mathbf{p}^s)
=
(1 - \gamma) \, \epsilon_\theta(\mathbf{p}^d_t, t, \mathbf{0}) + \gamma \, \epsilon_\theta(\mathbf{p}^d_t, t, \mathbf{p}^s),
\end{align}
where $\gamma > 0$ (we set $\gamma=6$, as in LiDiff), and $\epsilon_\theta$ is conditioned with $\mathbf{p}^s$ or with the \mbox{all-zero} point cloud $\mathbf{0}$. 

The network $\epsilon_\theta$ has the architecture proposed in LiDiff, but with all batch normalization layers replaced by instance normalization. %
Indeed, in LiDiff $\epsilon_\theta$ takes as input widely different point clouds for conditioning: $\mathbf{p}^s$ or $\mathbf{0}$. 
We noticed that the features statistics may vary significantly when passing $\mathbf{p}^s$ or $\mathbf{0}$, potentially corrupting the running statistics and leading to instabilities during inference. 
To avoid this issue, we use instance normalization layers in which the statistics of the features are computed for each instance during training and inference. We elaborate on this in~\cref{sec:exp_instancenorm}.

%% file: sections/04_results.tex
\section{Results}
\label{sec:results}

\para{Dataset.}
We train and evaluate our diffusion model on SemanticKITTI~\cite{semantickitti}, preparing ground-truth point clouds as in LiDiff~\cite{lidiff} by aggregating scans using the provided ego-poses and excluding moving objects. %
Training is done on sequences 00\hspace{1pt}-10, with sequence 08 reserved for validation.

\para{Implementation details.} 
We train our diffusion model for $40$ epochs using a batch size of 8 and Adam optimizer~\cite{adam} with a constant learning rate of $2\cdot10^{-4}$ and no weight decay.
The diffusion parameters are linearly interpolated between $\beta_1 = 3.5 \cdot 10^{-5}$ and $\beta_{1000} = 0.007$, as in LiDiff. 
The MinkUNet $\epsilon_\theta$~\cite{choy2019minkunet} (adapted as in LiDiff) is trained with and without conditioning, selected randomly per iteration, with conditioning used $10\%$ of the time. %
The voxel size of $\epsilon_\theta$ is $5$\,cm. 
The point clouds $\mathbf{p}^d$ and $\mathbf{p}^s$ are cropped 
at a range of $50$\,m. 
The sizes of $\mathbf{p}^s$ and $\mathbf{p}^d$ are limited to 18K and 180K points, with farthest point sampling and uniform sampling applied to $\mathbf{p}^s$ and $\mathbf{p}^d$, respectively.
The generation via DPM-Solver~\cite{dpmsolver} is done in 20 sampling steps.
Before subsampling and $K$-duplicating $\mathbf{p}^s$ to obtain $\tilde{\mathbf{p}}^s$, we add $1000$ points uniformly sampled from a flat disc of radius $3.5$\,m at the ground height to 
compensate for points close to the car removed during the data preprocessing.
Unless mentioned otherwise (with $\dagger$), the %
diffusion output is densified %
with the refinement network of LiDiff used off-the-shelf.

\para{Metrics.} 
For evaluation, we follow LiDiff.
We report Chamfer distance (CD), Jensen-Shannon divergence (JSD) computed in BEV and in 3D space, and occupancy IoU for different voxel sizes 
(0.5, 0.2 or 0.1~m).

\subsection{Comparison to existing methods}
\input{tables/completion_lidiff}

\input{tables/ablations}

\para{Diffusion only.} 
We first evaluate the quality of our learned global diffusion model, \method{}$^\dagger$. 
We compare it to the local diffusion model, LiDiff$^\dagger$, and to PVD \cite{pvd}, whose results on SemanticKITTI were obtained by LiDiff authors~\cite{lidiff}. 
No refinement is used after diffusion for any of these methods.
The results are reported in \cref{tab:benchmark}. We notice that \method{}$^\dagger$ performs better than the other diffusion-only methods on all metrics, except for the Chamfer distance (CD), where it is just $1$\,cm away from LiDiff$^\dagger$. 
We also remark that our method's IoU for voxels of size 0.5\,m and 0.2\,m is already better than the IoU obtained by LiDiff after refinement.

\para{Complete method.} 
We then compare to all baselines, still in \cref{tab:benchmark}.
We include the refinement step used for densification in LiDiff and \method{}.
We notice that \method{} outperforms all the others in terms of JSD BEV and IoU at 0.2\,m and 0.1\,m. \method{} is also better than LiDiff in JSD 3D and IoU at 0.5\,m, and reaches an equivalent CD. We only notice a slight performance drop for \method{} before and after refinement in JSD 3D and IoU at 0.5\,m. 
The qualitative results presented in \cref{fig:qualitative_results} show that the generated point clouds with \method{} exhibit less spurious structures than those generated with LiDiff, in particular far away from the center of the scenes.

\subsection{Hyperparameter study}
\label{sec:ablation}

\para{Reduced evaluation set.} 
For simplicity and faster evaluations, we conduct this study on a small subset of the validation sequence: 10 scans equally spaced in time.

\para{Start of the diffusion $t_0$.}
We generate point clouds using \cref{eq:denoising} for different starting points $t_0$, as defined in \cref{eq:start_our_diff}, and not using the DPM-Solver.
We show in \cref{ablations:ablations}(a) that the best results are observed for $t_0$ around $300$.
Starting the diffusion with $t_0 = 50$ prevents the network from completing the point cloud (low recall), as the starting point cloud $\tilde{\mathbf{p}}^d_{t_0}$ is not noisy enough to fill all the gaps present in $\mathbf{p}^s$. 
In contrast, picking $t_0$ close to $T$ improves the recall, but the original structures in $\mathbf{p}^s$ are less preserved (lower precision). 
Starting at $t_0=300$ is a sweet spot to optimize the Chamfer distance and the voxel IoU.
The conclusion remains the same after post-processing the results with the refinement network.

\input{figures/explosion}

\input{figures/qualitatives}

\para{DPM-Solver.}
For $t_0=300$, we study the achievable speedup to generate a point cloud by using the DPM-Solver. 
We test the number of steps to generate the point clouds, between $5$ and $50$ (see \cref{ablations:ablations}(b)). 
We notice that $20$ steps are enough while the results degrade when going below.

\subsection{Statistics of the predicted noise \texorpdfstring{$\epsilon$}{epsilon}}

In LiDiff, the authors notice that the predicted noise at the output of $\epsilon_\theta$ has a mean and standard deviation very far from the expected values of $0$ and $1$. They introduce regularization losses to enforce this behavior (see \cref{app:localdiff}\!\!). 
In our method, we notice that the network $\epsilon_\theta$ naturally predicts the noise with zero mean. 
Its standard deviation is only slightly underestimated (closer to $0.6$ than $1$).

\subsection{Preventing instabilities}
\label{sec:exp_instancenorm}

When training our network $\epsilon_\theta$ with batch normalization layers, we noticed instabilities where points were generated at very far ranges. 
We identified that it was due to the batch statistics, which could be quite different whether $\epsilon_\theta$ was conditioned or not. 
To avoid this issue, we replaced all batch normalization layers by instance normalization layers. 
The different behaviors are illustrated in \cref{fig:explosion}.

%% file: tables/completion_lidiff.tex
\begin{table}[t]
\caption{
\textbf{SemanticKITTI validation set.} Baselines, metrics and ground-truth data are from LiDiff~\cite{lidiff}\protect\footnotemark[1].
} %
\label{tab:benchmark}
\centering
\small
\resizebox{1.0\linewidth}{!}{%
\setlength{\tabcolsep}{3pt}
\begin{tabular}{l@{~}rcccccccccccc}
\toprule
\multicolumn{2}{l}{\multirow{2}{*}[-0.5mm]{Method}}
    & \multirow{2}{*}[-0.5mm]{\shortstack{Output}} 
    & \multirow{2}{*}[-0.5mm]{\shortstack{JSD\textdownarrow\\\hspace{-1mm}3D}}
    & \multirow{2}{*}[-0.5mm]{\shortstack{JSD\textdownarrow\\\hspace{-1mm}BEV}}
    & \multicolumn{3}{c}{Vox.~IoU\textuparrow} 
    & \multirow{2}{*}[-0.5mm]{CD\textdownarrow}
\\
\cmidrule(lr){6-8}
    & 
    &  
    &  
    & 
    & 0.5 
    & 0.2 
    & 0.1 
    & 
\\
\midrule
LMSCNet & \cite{lmscnet}
    & Voxel  & -     & 0.431 & 30.8 & 12.1 &  3.7 & 0.641 \\

\midrule
LODE    & \cite{li2023lode}
    & Surface & -     & 0.451 & 33.8 & 16.4 &  5.0 & 1.029 \\

MID     & \cite{vizzo2022makeitdense} 
    & Surface & -     & 0.470 & 31.6 & 22.7 & 13.1 & 0.503 \\

\midrule
PVD     & \cite{pvd}                    
    & Points   & -     & 0.498 & 15.9 &  4.0 &  0.6 & 1.256 \\

LiDiff$^\dagger$ & \cite{lidiff} & Points & 0.564 & 0.444 & 31.5 & 16.8 &  4.7 & 0.434 \\
\rowcolor{blue!15}
\multicolumn{2}{l}{\method{}$^\dagger$ (ours)} & Points & \bf 0.532 & 0.440 & 34.1 & 19.5 &  6.3 & 0.446\\

LiDiff  & \cite{lidiff} & Points   & 0.573 & 0.416 & 32,4 & 23.0 & 13.4 & \bf 0.376 \\
\rowcolor{blue!15}
\multicolumn{2}{l}{\method{}\;\, (ours)}                                
    & Points   & 0.542 & \bf 0.403 & \bf 36.6 & \bf 25.8 & \bf 14.9 & 0.377 \\
\bottomrule
\multicolumn{9}{l}{$^\dagger$: diffusion-only, \ie, without post-processing (refinement).} \\\
\end{tabular}}
\vspace{-3mm}
\end{table}

\footnotetext[1]{The Voxel IoU values differ from the camera-ready version as per the correction in \href{https://github.com/PRBonn/LiDiff/commit/e5305f93d1d9d9ad09fc033a6dc3408e061401c3}{LiDiff evaluation code}.}

%% file: tables/ablations.tex
\begin{table}[t!]
\caption{\textbf{Ablation studies.}\protect\footnotemark[1]}
\label{ablations:ablations}
\centering
\small

\resizebox{1.0\linewidth}{!}{
\setlength{\tabcolsep}{4pt}
\begin{tabular}{lcccccccc}
\multicolumn{9}{c}{\textbf{(a)} Effect of $t_0$, starting point of the diffusion.} \\
\toprule
    & \multicolumn{4}{c}{Diffusion only} 
    & \multicolumn{4}{c}{Diffusion + Refinement}
\\
\cmidrule(lr){2-5}
\cmidrule(lr){6-9}
\multirow{2}{*}{Start $t_0$}
    & \multirow{2}{*}{CD\textdownarrow} 
    & \multicolumn{3}{c}{Voxel~IoU\textuparrow} 
    & \multirow{2}{*}{CD\textdownarrow} 
    & \multicolumn{3}{c}{Voxel~IoU\textuparrow} 
\\
\cmidrule(lr){3-5}
\cmidrule(lr){7-9}
    &    
    & 0.5 
    & 0.2 
    & 0.1         
    &    
    & 0.5 
    & 0.2 
    & 0.1 
\\
\midrule
    1000 & 0.600 & 23.8 & 12.0 & 3.9 & 0.535 & 25.4 & 15.7 & 8.6\\
    500  & \bf 0.437 & 31.6 & 17.4 & 6.1 & \bf 0.370 & 34.6 & 22.8 & 12.8 \\
    \rowcolor{blue!15}
    300  & 0.439 & \bf 32.3 & \bf 18.9 & 6.9 & 0.375 & \bf 35.4 & \bf 24.4 & \bf 14.3 \\
    100  & 0.474 & 29.4 & 18.8 & 7.5 & 0.427 & 30.5 & 21.6 & 13.9 \\
    50   & 0.490 & 27.0 & 17.5 & \bf 7.6 & 0.440 & 28.7 & 19.9 & 12.8 \\
\bottomrule
\end{tabular}}

~\\{\tiny ~\\}

\setlength{\tabcolsep}{8pt}
\begin{tabular}{ccccc}
\multicolumn{5}{c}{\textbf{(b)} Effect of the number of DPM-Solver~\cite{dpmsolver} steps.} \\
\toprule
Steps & CD\textdownarrow & IoU 0.5 & IoU 0.2 & IoU 0.1\\

\midrule
50	& 0.437 & 32.9 & 19.2 & 6.8\\
\rowcolor{blue!15}
20	& \bf 0.428 & \bf 33.9 & \bf 19.6 & 6.5\\
10	& 0.443 & 31.0 & 18.0 & \bf 6.9\\
5	& 0.447 & 30.7 & 17.5 & 6.8\\
\bottomrule
\end{tabular}

\vspace{-5mm}
\end{table}

%% file: figures/explosion.tex
\begin{figure}[t]
    \centering
    \setlength{\tabcolsep}{1pt}
    \begin{tabular}{c|c}
        \noindent\includegraphics[width=0.48\linewidth, trim={0 0 0 0.7cm},clip]{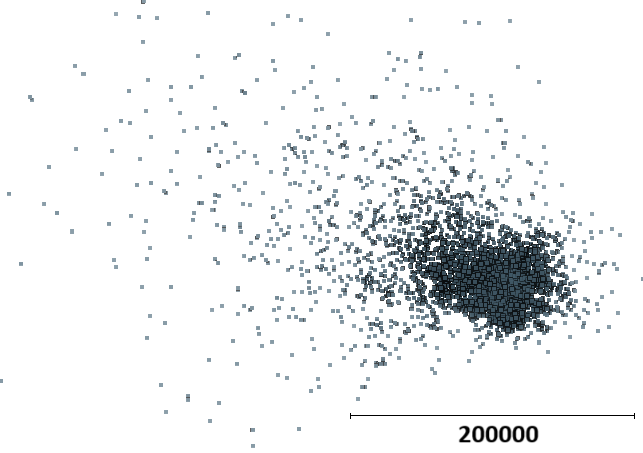}&
        \noindent\includegraphics[width=0.48\linewidth, trim={0 0 0 0.7cm},clip]{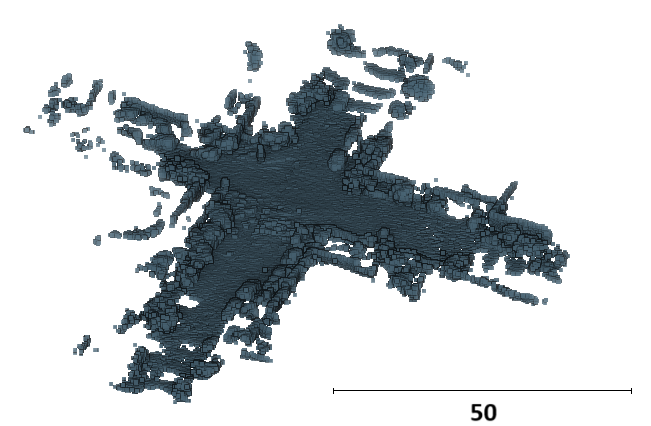}\\
        \small BatchNorm & \small InstanceNorm
    \end{tabular}
    \caption{\textbf{Effect of normalization.} Using batch normalization layers (left) leads to instabilities with points generated at far ranges. This is solved when using instance normalization layers instead (right). Note the scale difference. %
    }
    \label{fig:explosion}
\vspace*{-5mm}
\end{figure}

%% file: figures/qualitatives.tex
\begin{figure*}[t]
    \centering
    \small
    \setlength{\tabcolsep}{3pt}
    {%
    \begin{tabular}{cccc}
        LiDiff$^\dagger$ (no refinement) & \method{}$^\dagger$ (no refinement) & LiDiff & \method{} \\
        \includegraphics[width=0.23 \linewidth, height=0.14\linewidth]{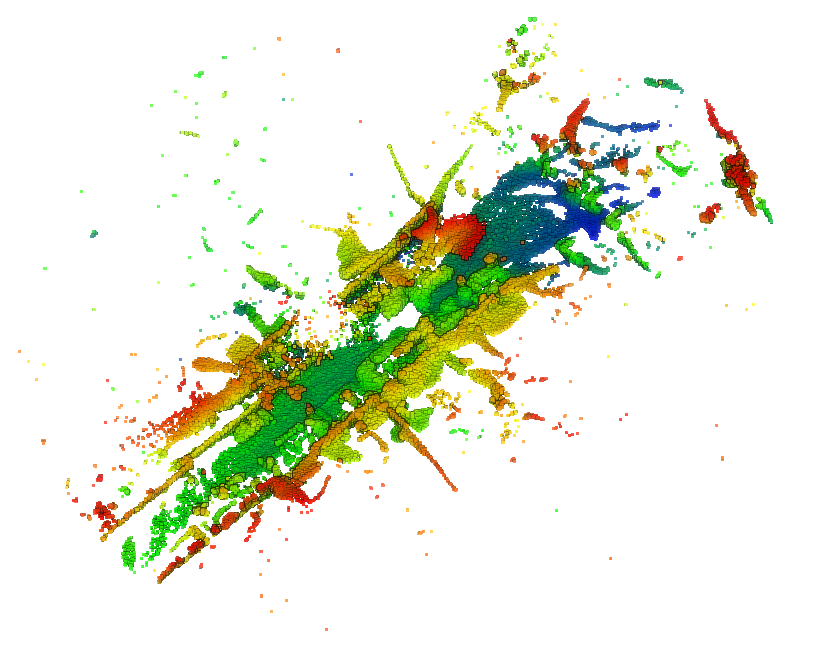} & 
        \includegraphics[width=0.23 \linewidth, height=0.14\linewidth]{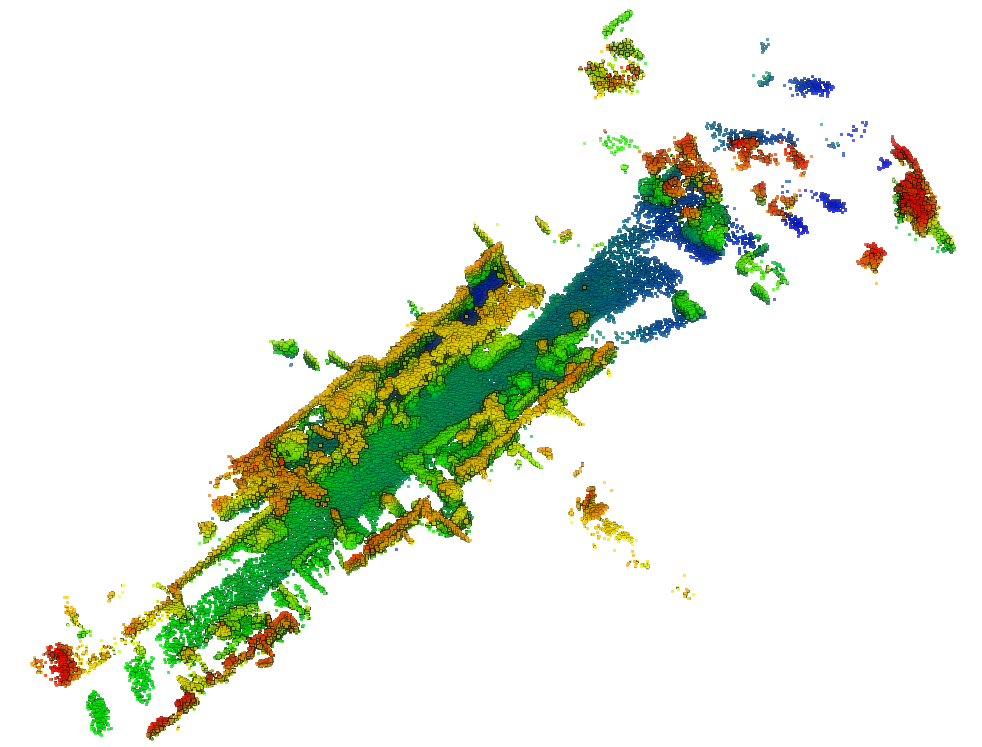} & 
        \includegraphics[width=0.23 \linewidth, height=0.14\linewidth]{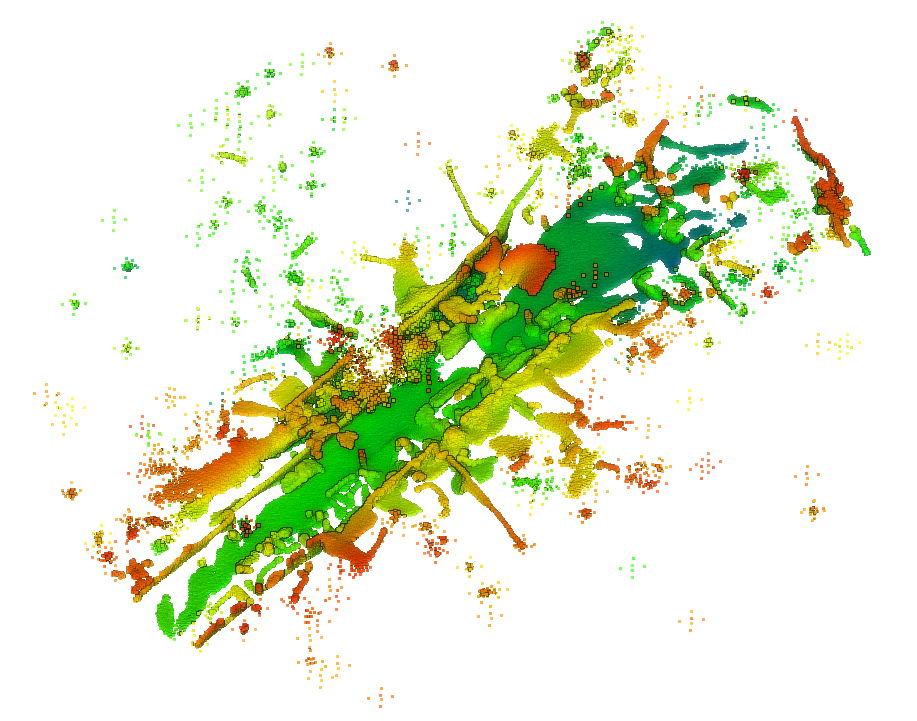} & 
        \includegraphics[width=0.23 \linewidth, height=0.14\linewidth]{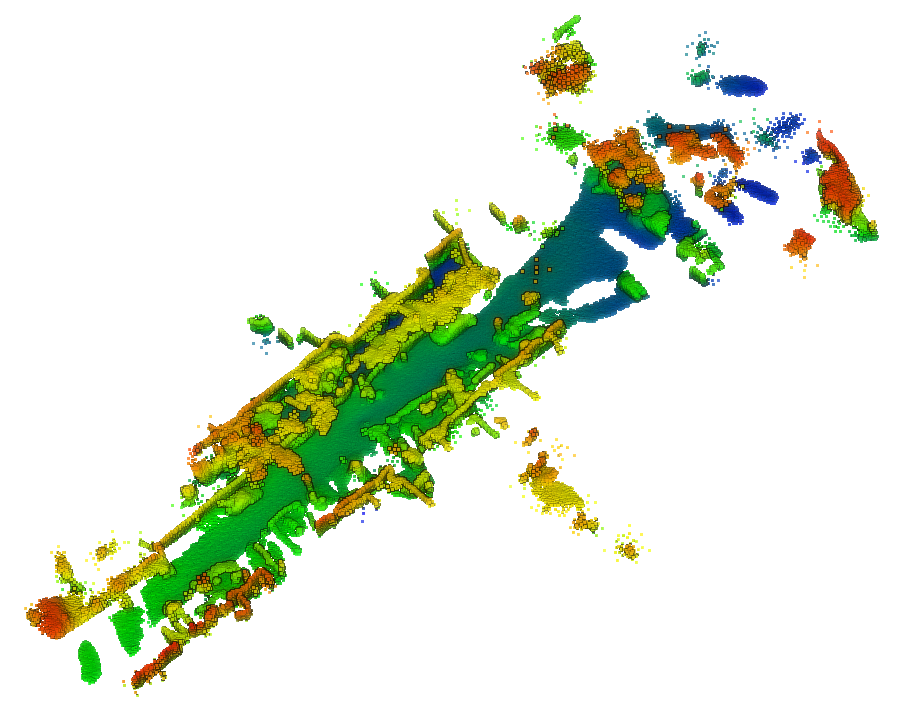} \\[-.5em]
        \includegraphics[width=0.23 \linewidth, height=0.14\linewidth]{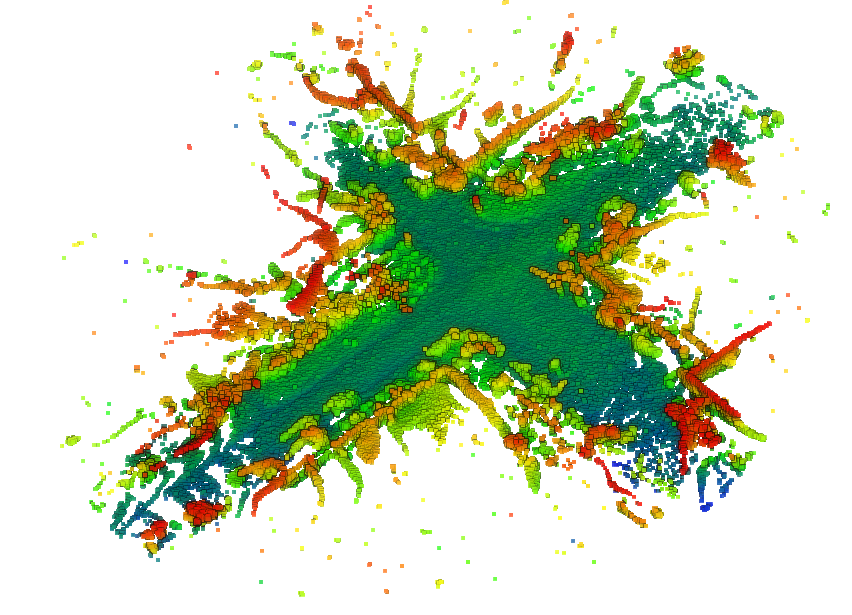} & 
        \includegraphics[width=0.23 \linewidth, height=0.14\linewidth]{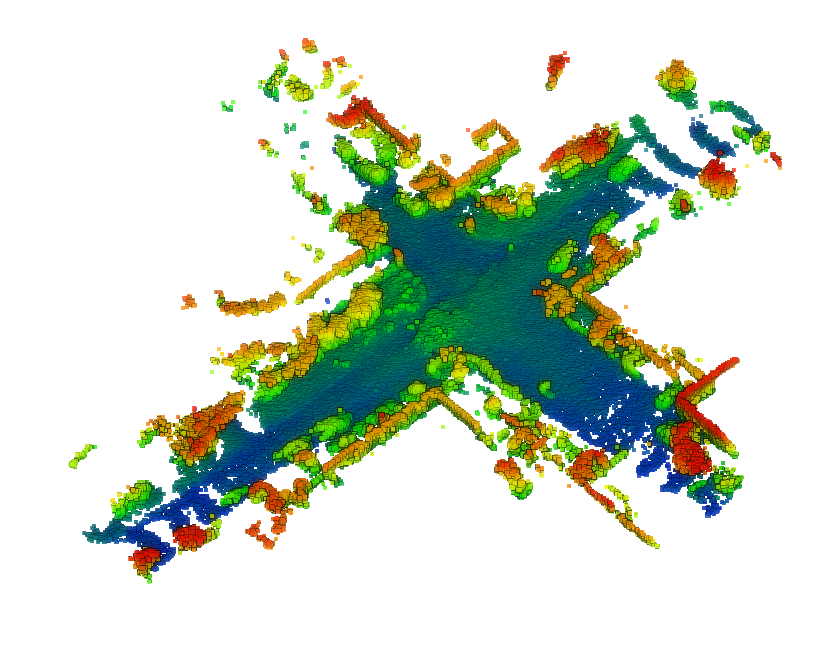} & 
        \includegraphics[width=0.23 \linewidth, height=0.14\linewidth]{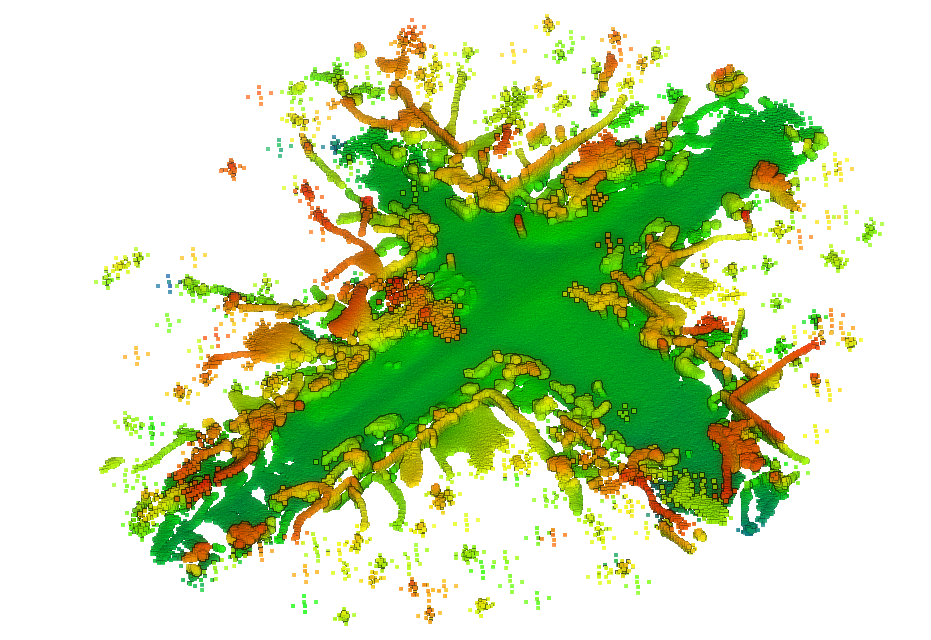} & 
        \includegraphics[width=0.23 \linewidth, height=0.14\linewidth]{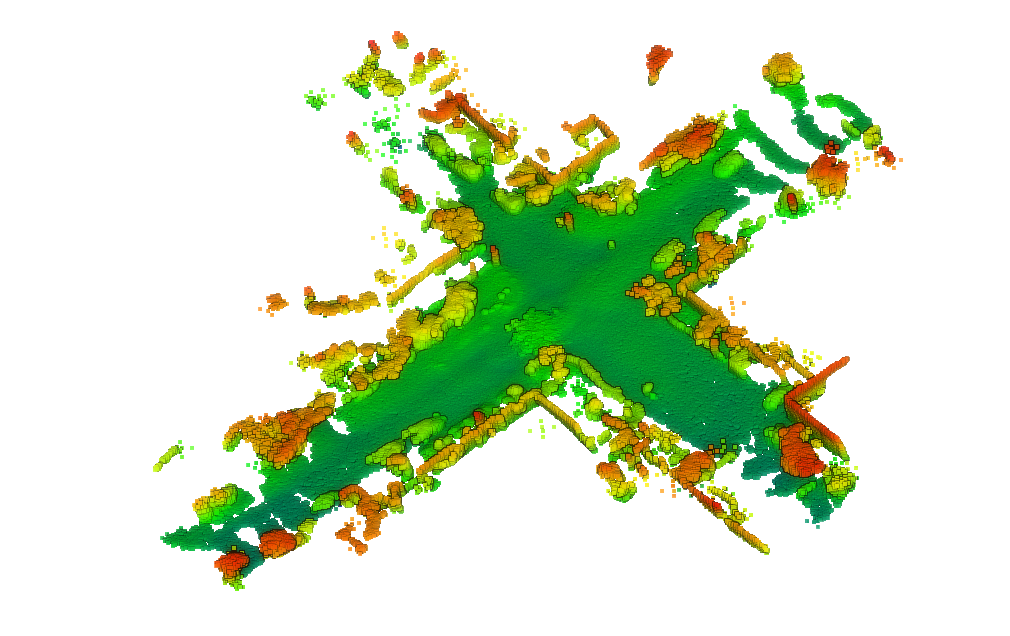} \\[-0.5em]
        \includegraphics[width=0.23 \linewidth, height=0.14\linewidth]{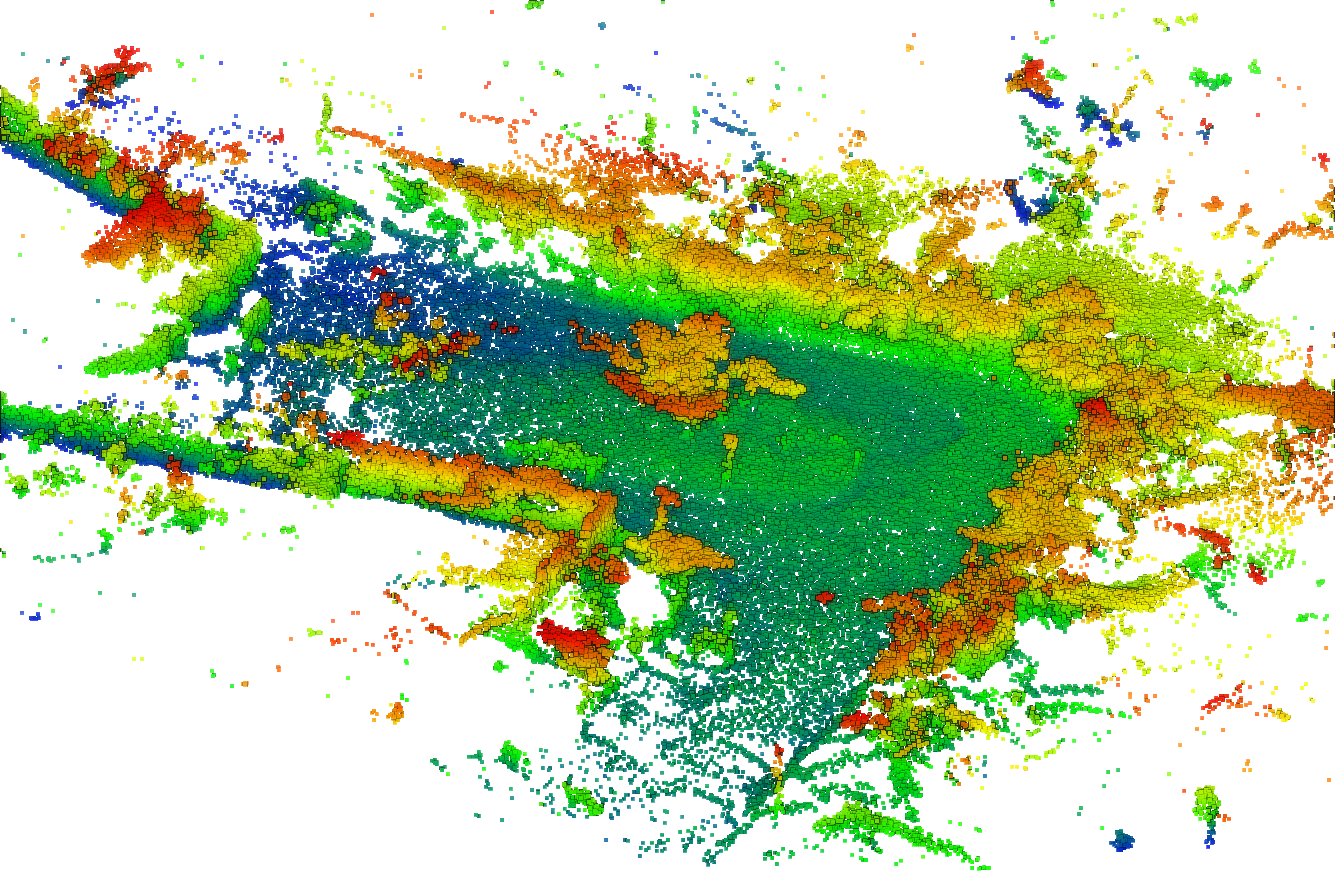} & 
        \includegraphics[width=0.23 \linewidth, height=0.14\linewidth]{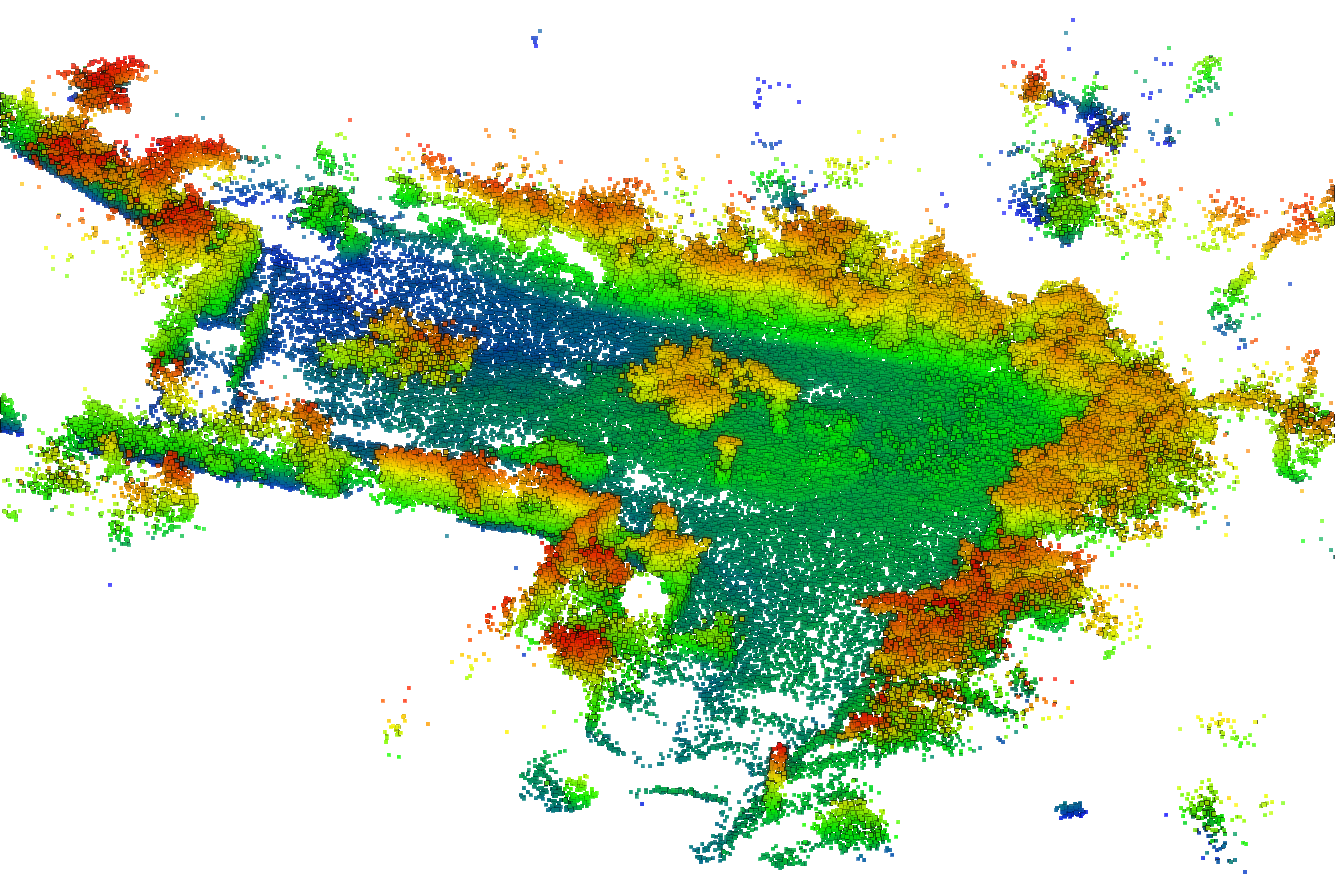} & 
        \includegraphics[width=0.23 \linewidth, height=0.14\linewidth]{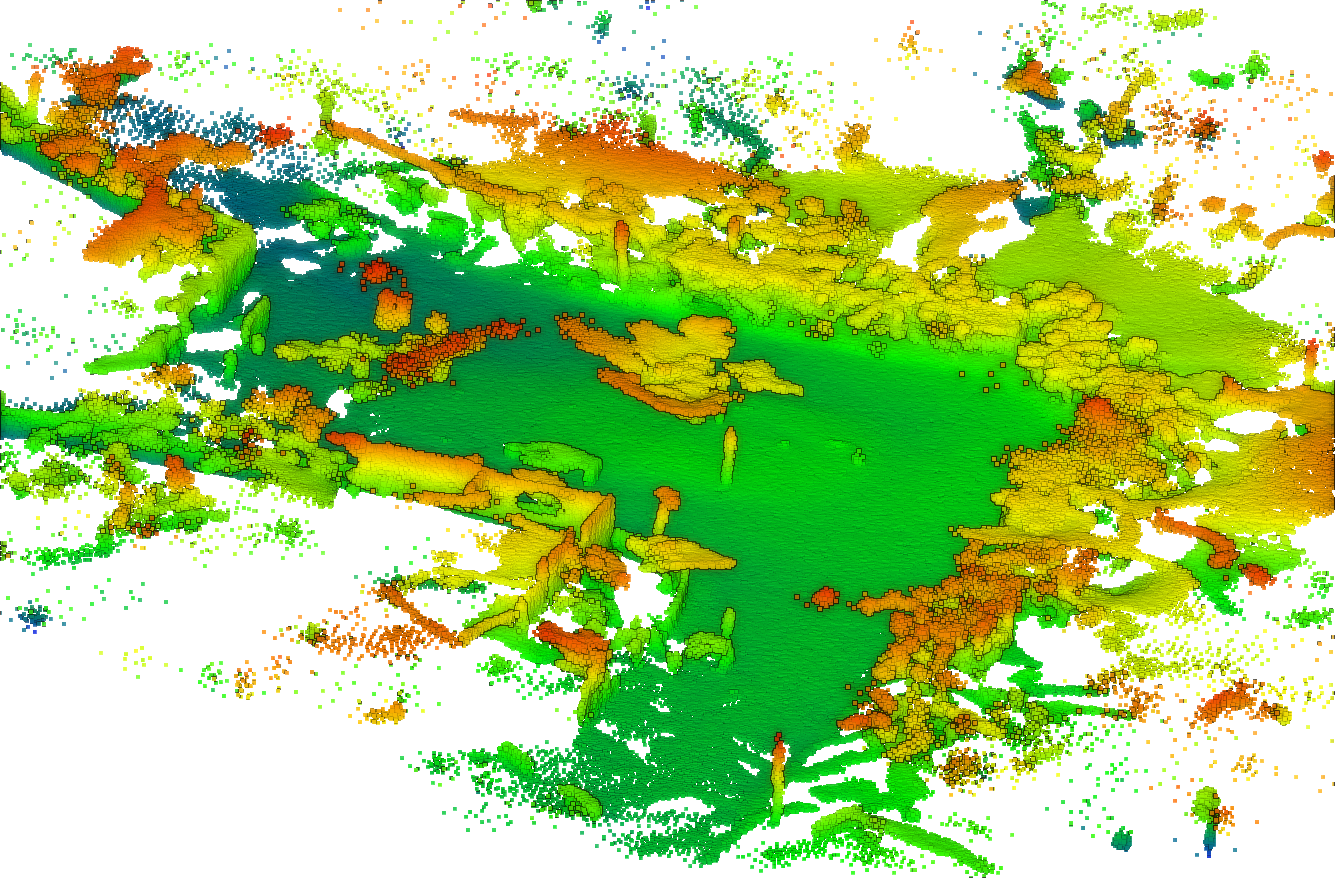} & 
        \includegraphics[width=0.23 \linewidth, height=0.14\linewidth]{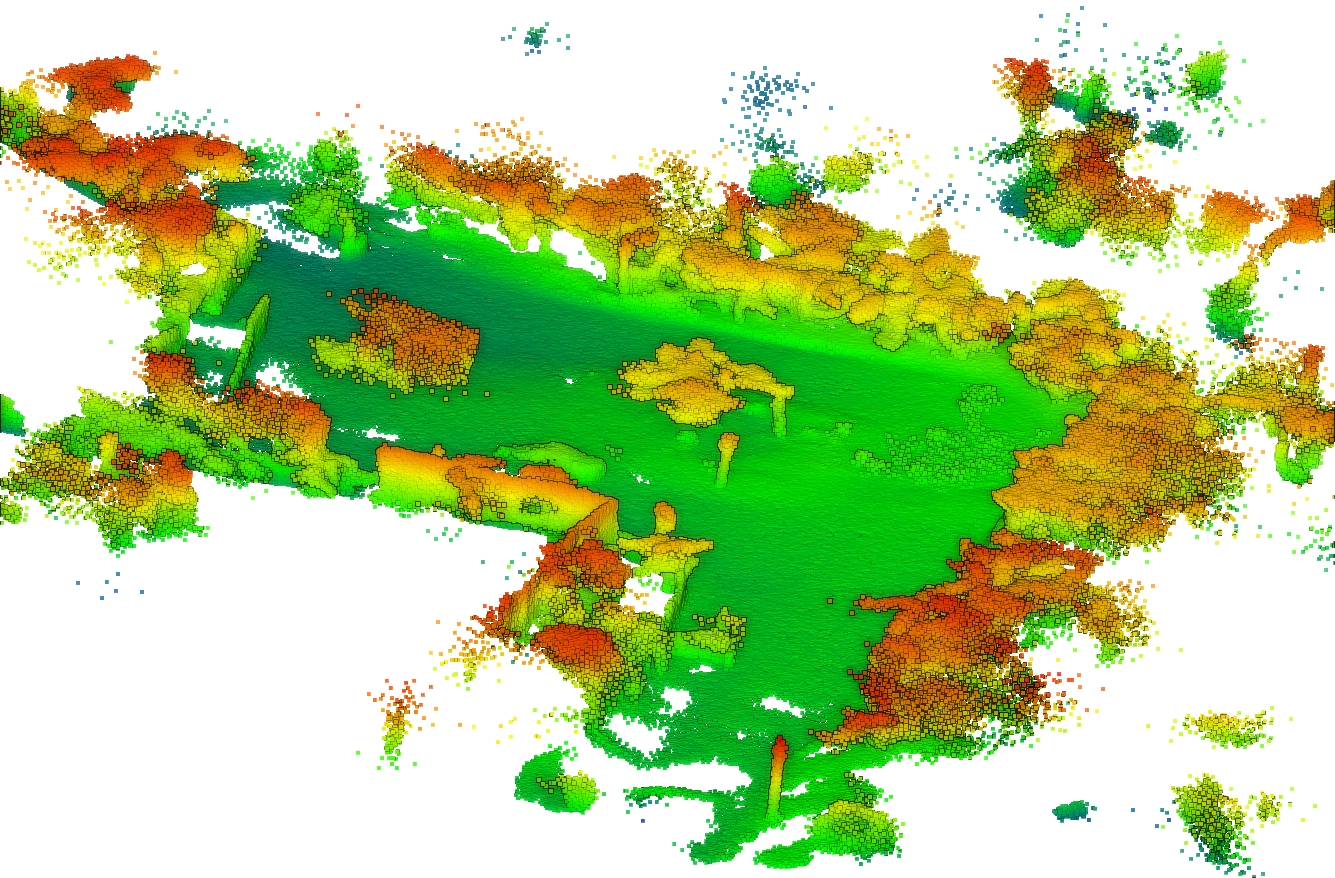} \\
    \end{tabular}%
    }
    \caption{\textbf{Qualitative results}. We show results from local diffusion of LiDiff w/o refinement (first column), after global diffusion at scene level with \method{} w/o refinement (second column), with LiDiff including refinement (third column), and with \method{} followed by refinement (last column).}%
    \label{fig:qualitative_results}
\vspace*{-3mm}
\end{figure*}

%% file: sections/05_conclusion.tex
\section{Conclusion}
\label{sec:conclusion}
We have shown that vanilla DDPM can be applied to complete outdoor lidar point clouds, meaning that local point diffusion, as used in prior work~\cite{lidiff,scorelidar}, is not necessary at the scene level. 
\method{} surpasses existing point-level diffusion-based methods on SemanticKITTI. 
In addition, by following \ddpm{} formulation, \method{} opens the door to unconditional generation, as presented in \cref{fig:generation}. 
This is achieved by deactivating the conditioning (setting $\gamma=0$ in \cref{eq:our_cfg}) and replacing the input scan $\tilde{\mathbf{p}}^s$ with $180K$-point clouds sampled from arbitrary shapes.  
\input{figures/generation}

%% file: figures/generation.tex
\begin{figure}[!ht]
    \centering
    \small
    \setlength{\tabcolsep}{3pt}
    \resizebox{\linewidth}{!}{%
    \begin{tabular}{ccc}
        \includegraphics[width=0.4 \linewidth]{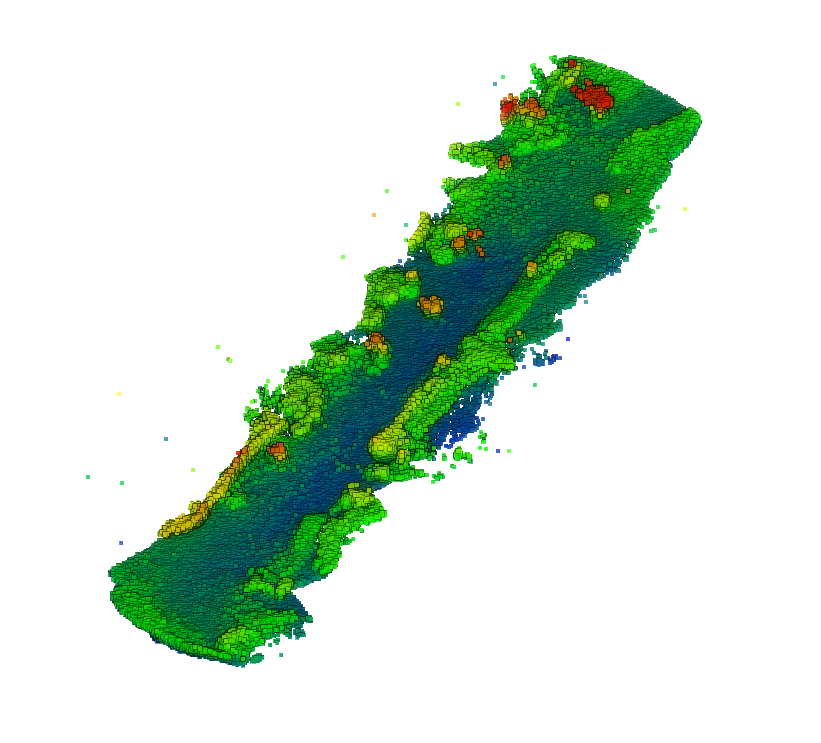} & 
        \includegraphics[width=0.4 \linewidth]{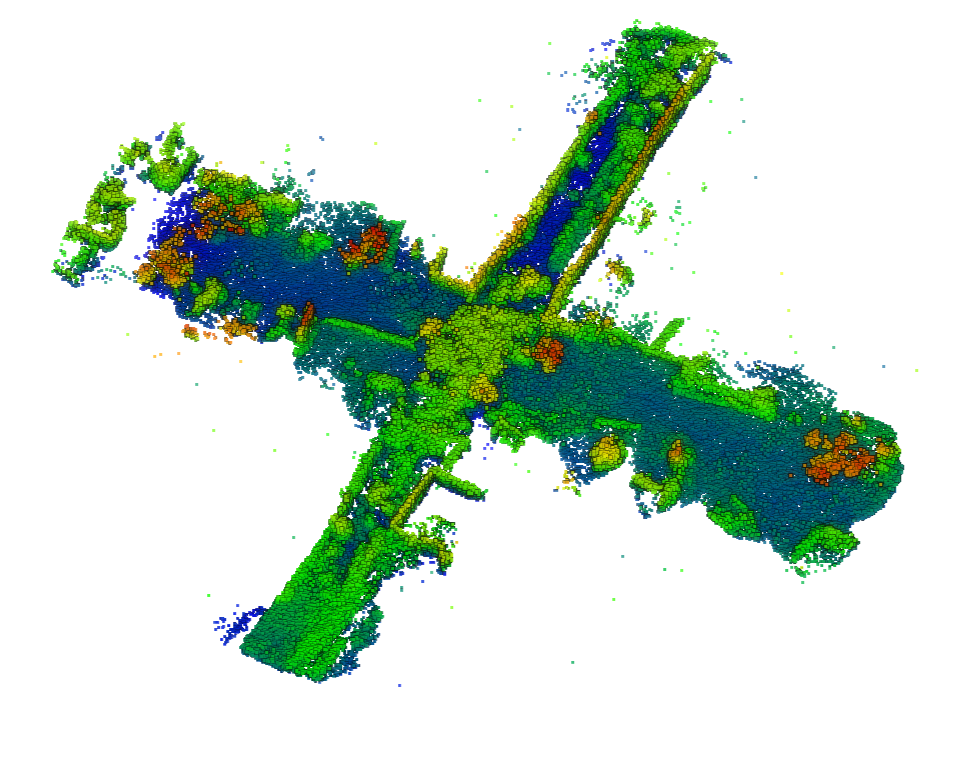} &
        \includegraphics[width=0.4 \linewidth]{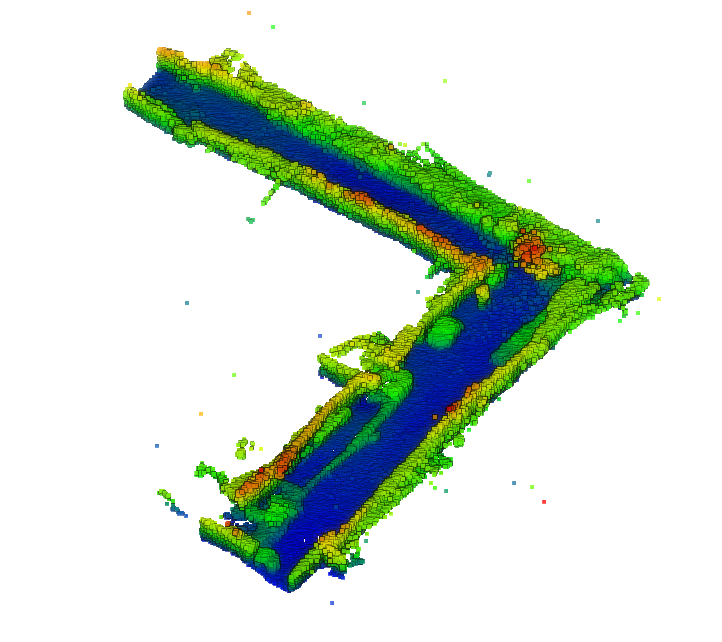} \\
    \end{tabular}%
    }
\vspace*{-5mm}
    \caption{\textbf{Pure generation with \method.} We generate scenes unconditionally, from arbitrary point clouds following a straight, crossing, or turn shape.}
    \label{fig:generation}
\vspace*{0mm}
\end{figure}

%% file: sections/0X_appendix.tex
\appendix
\section{Appendix}
\label{app:localdiff}

We present here the diffusion formulation as a \emph{local} problem, as discussed and used in prior work~\cite{lidiff,scorelidar}. 
We then highlight concerns that this formulation adds complexity and may introduce approximations in the denoising process.

\subsection{Formulation of the diffusion process as a local denoising}

{Starting from a dense point cloud $\mathbf{p}_0^d$, the forward process in LiDiff~\cite{lidiff} satisfies

\begin{align}
\label{eq:forward_lidiff}
\mathbf{p}_t^d = \mathbf{p}_0^d + \sqrt{1 - \bar{\alpha}_t} \; \mathbf{\epsilon},
\end{align}
which is obtained by {considering point offsets with respect to the ground truth $\mathbf{p}_0^d$,} setting 
\begin{align}
\label{eq:link_x_p}
\mathbf{x}_t = \mathbf{p}_t^d  - \mathbf{p}_0^d \quad \text{and}  \quad \mathbf{x}_0 = \mathbf{0} 
\end{align}
in \cref{eq:forward}. The endpoint of this forward local diffusion process, or, equivalently, the starting point of the reverse denoising process, $\mathbf{p}_T^d$, is thus a noisy version of $\mathbf{p}_0^d$.

Second, conditioning $\epsilon_\theta$ on $\mathbf{x}_t$ as in 
\cref{eq:learned_neural_net} is not possible as it would require knowing $\mathbf{p}_0^d$ during generation. Therefore, $\epsilon_\theta$ is instead conditioned on $\mathbf{p}_t^d$ and trained to minimize
\begin{align}
\label{eq:learned_neural_net_lidff}
& \mathcal{L}(\theta) 
= 
\mathbb{E}_{\mathbf{p}^d_t} \left[ \| \mathbf{\epsilon} - \epsilon_\theta(\mathbf{p}^d_t, t) \|^2 
+ \lambda \, \mathcal{L}_{\rm reg}(\epsilon_\theta(\mathbf{p}^d_t, t))\right],\\
& \text{with}\;
 \mathcal{L}_{\rm reg}(\epsilon_\theta(\mathbf{p}^d_t, t)) 
= \mathcal{L}_{\rm mean}(\epsilon_\theta(\mathbf{p}^d_t, t))
+ \mathcal{L}_{\rm std}(\epsilon_\theta(\mathbf{p}^d_t, t))\,,\nonumber
\end{align}
where $\lambda>0$, and $\mathcal{L}_{\rm mean},\mathcal{L}_{\rm std}$ guide the mean and standard deviation over $\epsilon_\theta(\mathbf{p}^d_t, t)$ to $0$ and $1$, respectively, as $\epsilon_\theta$ predicts peaky distribution, far from expected $\mathcal{N}(\mathbf{0}, \mathbf{I})$.

Third, using Eqs. (\ref{eq:link_x_p}) in (\ref{eq:denoising}), the reverse denoising becomes
\begin{align}
\label{eq:pdt}
\mathbf{p}_{t-1}^d 
= 
\mathbf{p}_{0}^d 
+ 
\frac{\mathbf{p}_{t}^d - \mathbf{p}_{0}^d}{\sqrt{\alpha_t}} 
- 
\frac{(1 - \alpha_t) \, \epsilon_\theta(\mathbf{p}^d_t, t)}{\sqrt{\alpha_t \, (1 - \bar{\alpha}_t)}}  
+ 
\sqrt{\beta_t} \; \mathbf{z}\,.
\end{align}
Yet, this formula is unusable in practice as it requires access to the dense ground-truth point cloud $\mathbf{p}_{0}^d$, which is unknown during generation. As a workaround, LiDiff replaces
the dense point cloud $\mathbf{p}_{0}^d$ by a noisy estimate $\tilde{\mathbf{p}}^s$, %
obtained by noising $K$ duplications of the sparse point cloud $\mathbf{p}^s$. 
Ultimately, the approximate reverse denoising process writes
\begin{align}
\label{eq:denoising_lidiff}
\mathbf{p}_{t-1}^d 
= 
\tilde{\mathbf{p}}^s
+ 
\frac{\mathbf{p}_{t}^d - \tilde{\mathbf{p}}^s}{\sqrt{\alpha_t}} 
- 
\frac{(1 - \alpha_t) \, \epsilon_\theta(\mathbf{p}^d_t, t)}{\sqrt{\alpha_t \, (1 - \bar{\alpha}_t)}}  
+ 
\sqrt{\beta_t} \; \mathbf{z}\,.
\end{align}

Finally, one needs to construct a starting point $\mathbf{p}_T^d$ of the reverse denoising process. 
According to the forward process in \cref{eq:forward_lidiff}, a starting point can be constructed by sampling $\mathbf{z} \sim \mathcal{N}(\mathbf{0}, \mathbf{I})$ and adding it to $\mathbf{p}_0^d$, \ie, 
\begin{align}
\mathbf{p}_T^d 
= \mathbf{p}_0^d + \sqrt{1 - \bar{\alpha}_T} \; \mathbf{z}
\,\simeq\, \mathbf{p}_0^d + \mathbf{z},
\end{align}
where we use the fact that $\sqrt{1 - \bar{\alpha}_T} \simeq 1$, as in vanilla DDPM. Again, it would require knowing the ground truth $\mathbf{p}_0^d$. Instead, the starting point is also approximated  using $\tilde{\mathbf{p}}^s$:
\begin{align}
\label{eq:start_lidiff}
\mathbf{p}_T^d = \tilde{\mathbf{p}}^s 
+ \mathbf{z}, \quad \text{ where } \quad \mathbf{z} \sim \mathcal{N}(\mathbf{0}, \mathbf{I}).
\end{align}
}

\subsection{Limitations of local denoising paradigm}

While the above local point diffusion proved to work, it relies on major approximations, which we outline here.

\textbf{a.~}The justification of LiDiff to propose local point diffusion is that DDPM requires normalization, which leads to compression of data along some of the point cloud axes (Sec.~3.3 of LiDiff). 
We believe this observation may be flawed.
Not only could normalization be applied within fixed bounds, but it is actually not needed for DDPM, which can be formulated in the metric space (see~\cref{sec:meth_meth}).

\textbf{b.~}Diffusion requires the predicted noise to follow a standard normal distribution. 
To compensate for the predicted local noise deviating from a zero-mean, unit-variance Gaussian, the authors add regularization losses (Sec.~3.4 of LiDiff), which further complicates the optimization objective.

\textbf{c.~}A key limitation of local diffusion is the reliance on a noisy estimate $\tilde{\mathbf{p}}^s$ of the ground truth in \cref{eq:pdt} and \cref{eq:start_lidiff}, which hinders its use for data generation, thereby losing one of the central benefits of diffusion models.

In this work, we show that DDPM can be applied to non-normalized point clouds without losing detail; hence, the local (re-)formulation is not required~(\textbf{a.}).
Moreover, our vanilla scene-level formulation avoids the need for added regularization~(\textbf{b.}). 
Last, a byproduct of our choices opens up the possibility to generate point clouds~(\textbf{c.}).

\section*{Acknowledgment}

This work has been carried out using HPC resources from GENCI-IDRIS (grants AD011014484R1, AD011012883R3).
We thank Mickael Chen, Corentin Sautier and Mohammad Fahes for the proofreading and valuable feedback.

%% file: main.bbl
\begin{thebibliography}{10}
\providecommand{\url}[1]{#1}
\csname url@rmstyle\endcsname
\providecommand{\newblock}{\relax}
\providecommand{\bibinfo}[2]{#2}
\providecommand\BIBentrySTDinterwordspacing{\spaceskip=0pt\relax}
\providecommand\BIBentryALTinterwordstretchfactor{4}
\providecommand\BIBentryALTinterwordspacing{\spaceskip=\fontdimen2\font plus
\BIBentryALTinterwordstretchfactor\fontdimen3\font minus \fontdimen4\font\relax}
\providecommand\BIBforeignlanguage[2]{{%
\expandafter\ifx\csname l@#1\endcsname\relax
\typeout{** WARNING: IEEEtran.bst: No hyphenation pattern has been}%
\typeout{** loaded for the language `#1'. Using the pattern for}%
\typeout{** the default language instead.}%
\else
\language=\csname l@#1\endcsname
\fi
#2}}

\bibitem{popovic2021volumetric}
M.~Popović, F.~Thomas, S.~Papatheodorou, N.~Funk, T.~Vidal-Calleja, and S.~Leutenegger, ``Volumetric occupancy mapping with probabilistic depth completion for robotic navigation,'' \emph{RA-L}, 2021.

\bibitem{vizzo2022makeitdense}
I.~Vizzo, B.~Mersch, R.~Marcuzzi, L.~Wiesmann, J.~Behley, and C.~Stachniss, ``Make it dense: Self-supervised geometric scan completion of sparse {3D} lidar scans in large outdoor environments,'' \emph{RA-L}, 2022.

\bibitem{wu2022sparse}
X.~Wu, L.~Peng, H.~Yang, L.~Xie, C.~Huang, C.~Deng, H.~Liu, and D.~Cai, ``Sparse fuse dense: Towards high quality {3D} detection with depth completion,'' in \emph{CVPR}, 2022.

\bibitem{xiong2023ultralidar}
Y.~Xiong, W.-C. Ma, J.~Wang, and R.~Urtasun, ``Learning compact representations for lidar completion and generation,'' in \emph{CVPR}, 2023.

\bibitem{shan2023scp}
Y.~Shan, Y.~Xia, Y.~Chen, and D.~Cremers, ``Scp: Scene completion pre-training for 3d object detection,'' in \emph{ISPRS}, 2023.

\bibitem{michele2024saluda}
B.~Michele, A.~Boulch, G.~Puy, T.-H. Vu, R.~Marlet, and N.~Courty, ``{SALUDA}: Surface-based automotive lidar unsupervised domain adaptation,'' in \emph{3DV}, 2024.

\bibitem{sanchez2023domaingeneralization}
J.~Sanchez, J.-E. Deschaud, and F.~Goulette, ``Domain generalization of {3D} semantic segmentation in autonomous driving,'' in \emph{ICCV}, 2023.

\bibitem{yi2021completeandlabel}
L.~Yi, B.~Gong, and T.~Funkhouser, ``Complete \& {Label}: A domain adaptation approach to semantic segmentation of lidar point clouds,'' in \emph{CVPR}, 2021.

\bibitem{lidiff}
L.~Nunes, R.~Marcuzzi, B.~Mersch, J.~Behley, and C.~Stachniss, ``Scaling diffusion models to real-world 3d lidar scene completion,'' in \emph{CVPR}, 2024.

\bibitem{ddpm}
J.~Ho, A.~Jain, and P.~Abbeel, ``Denoising diffusion probabilistic models,'' in \emph{NeurIPS}, 2020.

\bibitem{pvd}
L.~Zhou, Y.~Du, and J.~Wu, ``3d shape generation and completion through point-voxel diffusion,'' in \emph{ICCV}, 2021.

\bibitem{dit3d}
S.~Mo, E.~Xie, R.~Chu, L.~Hong, M.~Niessner, and Z.~Li, ``Dit-3d: Exploring plain diffusion transformers for 3d shape generation,'' in \emph{NeurIPS}, 2023.

\bibitem{semantickitti}
J.~Behley, M.~Garbade, A.~Milioto, J.~Quenzel, S.~Behnke, C.~Stachniss, and J.~Gall, ``Semantickitti: A dataset for semantic scene understanding of lidar sequences,'' in \emph{CVPR}, 2019.

\bibitem{luo2021diffusion}
S.~Luo and W.~Hu, ``Diffusion probabilistic models for 3d point cloud generation,'' in \emph{CVPR}, 2021.

\bibitem{ma2018sparse}
F.~Ma and S.~Karaman, ``Sparse-to-dense: Depth prediction from sparse depth samples and a single image,'' in \emph{ICRA}, 2018.

\bibitem{jaritz2018sparse}
M.~Jaritz, R.~De~Charette, E.~Wirbel, X.~Perrotton, and F.~Nashashibi, ``Sparse and dense data with cnns: Depth completion and semantic segmentation,'' in \emph{3DV}, 2018.

\bibitem{lorensen1987marchingcubes}
W.~E. Lorensen and H.~E. Cline, ``Marching cubes: A high resolution {3D} surface construction algorithm,'' in \emph{SIGGRAPH}, 1987.

\bibitem{li2023lode}
P.~Li, R.~Zhao, Y.~Shi, H.~Zhao, J.~Yuan, G.~Zhou, and Y.-Q. Zhang, ``{LODE}: Locally conditioned eikonal implicit scene completion from sparse lidar,'' in \emph{ICRA}, 2023.

\bibitem{roldao2020lmscnet}
L.~Rold{\~a}o, R.~de~Charette, and A.~Verroust-Blondet, ``{LMSCNet}: Lightweight multiscale {3D} semantic completion,'' in \emph{3DV}, 2020.

\bibitem{scpnet}
Z.~Xia, Y.~Liu, X.~Li, X.~Zhu, Y.~Ma, Y.~Li, Y.~Hou, and Y.~Qiao, ``Scpnet: Semantic scene completion on point cloud,'' in \emph{CVPR}, 2023.

\bibitem{sscnet}
S.~Song, F.~Yu, A.~Zeng, A.~X. Chang, M.~Savva, and T.~Funkhouser, ``Semantic scene completion from a single depth image,'' in \emph{CVPR}, 2017.

\bibitem{yan2021sparse}
X.~Yan, J.~Gao, J.~Li, R.~Zhang, Z.~Li, R.~Huang, and S.~Cui, ``Sparse single sweep lidar point cloud segmentation via learning contextual shape priors from scene completion,'' in \emph{AAAI}, 2021.

\bibitem{roldao20223d}
L.~Roldao, R.~De~Charette, and A.~Verroust-Blondet, ``3d semantic scene completion: A survey,'' \emph{IJCV}, 2022.

\bibitem{rist2022semscenecomp}
C.~B. Rist, D.~Emmerichs, M.~Enzweiler, and D.~M. Gavrila, ``Semantic scene completion using local deep implicit functions on lidar data,'' \emph{T-PAMI}, 2022.

\bibitem{cai2020shapegf}
R.~Cai, G.~Yang, H.~Averbuch-Elor, Z.~Hao, S.~Belongie, N.~Snavely, and B.~Hariharan, ``Learning gradient fields for shape generation,'' in \emph{ECCV}, 2020.

\bibitem{yuan2018pcn}
W.~Yuan, T.~Khot, D.~Held, C.~Mertz, and M.~Hebert, ``Pcn: Point completion network,'' in \emph{3DV}, 2018.

\bibitem{boulch2021needrop}
A.~Boulch, P.-A. Langlois, G.~Puy, and R.~Marlet, ``{NeeDrop}: Self-supervised shape representation from sparse point clouds using needle dropping,'' in \emph{3DV}, 2021.

\bibitem{sulzer2022deep}
R.~Sulzer, L.~Landrieu, A.~Boulch, R.~Marlet, and B.~Vallet, ``Deep surface reconstruction from point clouds with visibility information,'' in \emph{ICPR}, 2022.

\bibitem{vizzo2021poisson}
I.~Vizzo, X.~Chen, N.~Chebrolu, J.~Behley, and C.~Stachniss, ``Poisson surface reconstruction for lidar odometry and mapping,'' in \emph{ICRA}, 2021.

\bibitem{huang2023nksr}
J.~Huang, Z.~Gojcic, M.~Atzmon, O.~Litany, S.~Fidler, and F.~Williams, ``Neural kernel surface reconstruction,'' in \emph{CVPR}, 2023.

\bibitem{scorelidar}
S.~Zhang, A.~Zhao, L.~Yang, Z.~Li, C.~Meng, H.~Xu, T.~Chen, A.~Wei, P.~P. GU, and L.~Sun, ``Distilling diffusion models to efficient 3d lidar scene completion,'' \emph{arXiv:2412.03515}, 2024.

\bibitem{diffssc}
H.~Cao and S.~Behnke, ``{DiffSSC}: Semantic lidar scan completion using denoising diffusion probabilistic models,'' \emph{arXiv:2409.18092}, 2024.

\bibitem{classifierfree}
J.~Ho and T.~Salimans, ``Classifier-free diffusion guidance,'' in \emph{NeurIPS Workshop}, 2021.

\bibitem{adam}
D.~P. Kingma and J.~Ba, ``Adam: A method for stochastic optimization,'' in \emph{ICLR}, 2015.

\bibitem{choy2019minkunet}
C.~Choy, J.~Gwak, and S.~Savarese, ``{4D} spatio-temporal {ConvNets}: {Minkowski} convolutional neural networks,'' in \emph{CVPR}, 2019.

\bibitem{dpmsolver}
C.~Lu, Y.~Zhou, F.~Bao, J.~Chen, C.~Li, and J.~Zhu, ``{DPM-Solver}: A fast {ODE} solver for diffusion probabilistic model sampling in around 10 steps,'' in \emph{NeurIPS}, 2022.

\bibitem{lmscnet}
L.~Roldao, R.~de~Charette, and A.~Verroust-Blondet, ``Lmscnet: Lightweight multiscale 3d semantic completion,'' in \emph{3DV}, 2020.

\end{thebibliography}
